# How do media talk about the Covid-19 pandemic? Metaphorical thematic clustering in Italian online newspapers[1]

Lucia Busso & Ottavia Tordini

Aston University, Birmingham (UK)/ Università di Pisa, Pisa (Italy)

**Abstract:** The contribution presents a study on figurative language of the first months of the COVID-19 crisis in Italian online newspapers. Particularly, we contrast topics and metaphorical language used by journalists in the first and second phase of the government response to the pandemic in Spring 2020. The analysis is conducted on a journalistic corpus collected between February 24th and June 3rd, 2020, and is performed using both quantitative and qualitative approaches: Structural Topic Modelling (Roberts et al., 2016), Conceptual Metaphor Theory (Lakoff and Johnson, 1980), and qualitative-corpus based metaphor analysis (Charteris-Black, 2004). We find a significant shift in topics discussed across Phase 1 and Phase 2, and interesting overlaps in topic-specific metaphors. Using qualitative corpus analysis, we further discuss metaphorical collocations of the topics of Economy and Society.

## 1. Introduction

The first few months of 2020 will be remembered in history for the SARS-CoV-2 (or Covid-19) outbreak. The disease first started out in the Hubei province in China, and progressively spread throughout the world, leading the WHO to declare it a global pandemic on 11 March. Most of the world's countries have responded to the virus outbreak implementing lockdowns and quarantine measures, which deeply affected global and local economies, societies, and the psychological state of millions of people.

---

[1] Although the study is the result of a joint collaboration, LB is responsible for sections 2, 3, 4. OT is responsible for sections 5 and 6. Section 1 was written and revised by both authors.

Italy was the first western country to be hit by the Covid-19 virus and to enforce local and then national lockdown to tackle the emergency. On 24 February, the Italian government established drastic containment measures to manage the emergency in Lombardy, which were then extended to the entire country on March 9. These two dates signal the beginning of the so-called *Fase 1* ('phase 1') of Italy's response to the pandemic. This phase is characterised by complete lockdown: schools and non-essential workplaces are shut down, people are forbidden from leaving the house unless for essential reasons (grocery shopping, health emergencies, work). During these first weeks, the health crisis is at its peak: with thousands of new cases and hundreds of deaths each day, Italy is one of the worst-hit countries in the world. With the slow but steady decrease in cases and in mortality during Spring, the government announced on 27 April the partial and gradual reopening of the country – from 4 May – initiating the so-called *Fase 2* ('phase 2'). Lockdown measures were reduced, and slowly restaurants, shops and small businesses reopened to the public. This second phase culminated on 3 June, when the national borders were reopened and internal circulation among regions restored.

Although the Covid crisis persisted in Italy and in the world for many more months after June 2020, the first months – marked by the two phases decided by the government – are a fertile ground to investigate which metaphorical representations emerged in the early days of the pandemic. The use of metaphors to describe the Covid crisis is in fact a widespread and cross-cultural phenomenon, and metaphorical language is ubiquitous in the discourse around the pandemic: a simple Google search for "metaphors of coronavirus" returns almost 15M hits (as of 26 October 2020). The metaphorical language used in discourses of Covid has been at the centre of a rapidly growing body of literature, especially focused on the ILLNESS AS A WAR metaphor (see Benzi and Novarese, 2022 for an in-depth review). Scholars have variously recognised that mediatic discourse of the pandemic draws from a pool of conventional metaphors used for previous crises, and combines them in novel ways (Semino, 2021). The massive use of metaphors is not surprising, and has been studied in mediatic and public discourse is widespread for many structural crises (among others Nünning, 2012; Cardini, 2014; Huang and Holmgreen, 2020). Notably, the use of conventional and novel metaphors is used in public discourse to frame complex issues in terms of other more concrete ones (Hellsten, 2000). This process helps a faster and more "digestible" understanding of an issue, and often is intended to elicit an emotional response out of the audience (Citron and Goldberg, 2014) and facilitates persuasion (Peclová and Lu, 2018). However, it is also important to note that the use of metaphors in public discourse does not contribute to a "neutral way of perceiving and representing reality" (Semino, 2021, p. 51), as using metaphorical framings can facilitate certain inferences and hinder others. In fact, the use of a specific source domain to represent a target domain necessarily highlights certain aspects but backgrounds others that do not fit with the metaphorical representation (Maasen and Weingart, 1995; Kennedy, 2000). For example, a lot of the literature on Covid has looked at the ILLNESS IS A WAR metaphor (*inter alia* Bagli 2021; Connelly 2020; Martinez-Brawley and Gualda 2020; Sabucedo, Alzate and Hur 2020; Wicke and Bolognesi 2020, 2021), which scholars recognise to foreground a need for swift action and complete extermination, while the possibility of gradually adapting to living with the disease is backgrounded (Semino, 2021).

In the present paper we focus on the conventional and novel metaphors that are used by Italian online media outlets (newspapers, magazines, blogs) around the central themes discussed in the first months of the Covid-19 emergency. Capitalising on previous research on the topic (Busso and Tordini, 2021), we take a different approach to the one used in most of the recent literature that focus on metaphors of Covid as a disease. Since the Covid-19 pandemic represented both a sanitary and socio-economic crisis at the same time, touching every aspect of our societal and economic systems, we analyse figurative language of the Covid-19 pandemic that encompasses not only the sanitary aspect of the crisis, but also its socio-economic nature.

Metaphors of health and socio-economic crises are extremely well-studied in the literature, and different metaphors have been isolated. Particularly, scholars have recognised in socio-economic crises from the 19th century onwards a number of recurring conventional metaphors that use a variety of source domains: living beings, weather catastrophes, infectious diseases, war, movement, liquid, buildings, machines (see *inter alia* Nerghes et al., 2015; Arrese and Vara-Miguel, 2016; Besomi, 2019).

Metaphors of health crises and epidemics have also received significant attention. Representations of many outbreaks have been studied in the last decades, from AIDS to cancer, from SARS and H1N1 to Ebola epidemics (see e.g. Sontag, 1990; Wallis and Nerlich, 2005; Larson et al., 2005; Camus, 2009; Angeli, 2012; Demjén and Semino, 2016; Trčková, 2015; Balteiro, 2017). These studies found that metaphors of illnesses and viruses often rely on source domains of combat, natural disasters, and movement and transport.

The study adopts an integrated methodology that combines quantitative and qualitative analyses. Specifically, we adopt a Structural Topic Modelling algorithm (Roberts et al., 2016) to explore the thematic structure of our corpus. We then use corpus methods to retrieve conceptual metaphors specific to the thematic clusters, and we investigate to what extent metaphors in the Italian press evolve during the two phases using corpus-assisted content analysis. As Burgers (2016, p. 250) states, in fact, "the ways metaphors change can (...) reveal how conceptualizations of social topics change over time". We expect to see both metaphors conventionally used to conceptualise health and socio-economic crises, and novel metaphors emerged in the current situation, as "new metaphors (...) can have the power to define reality" (Lakoff and Johnson, 1980, p. 133).

More specifically, we answer the following research questions:

1. To what extent and how has the shift between Phase 1 and Phase 2 mutated the mediatic representation of the crisis?

2. What metaphors (conventional and novel) emerge from the data to discuss the different aspects of the Covid-19 crisis?

The paper is structured as follows. Section 2 illustrates corpus design and methodology. Section 3 describes the computational topic modelling analysis. In Section 4 we focus on the metaphors specific to each topic. Sections 5 and 6 present an in-depth qualitative analysis of the metaphorical networks of socio-economic topics. In Section 7 we summarise our findings and their significance.

**2. Materials and methods**

An ad-hoc corpus of journalistic texts was compiled during the two phases (24 February – 3 June 2020). Data collection was carried out integrating semi-automatic and manual methodologies. Specifically, the documents were retrieved through advanced searches in Google and with the *BootCat* web-scraping toolkit (Baroni and Bernardini, 2005). Searches were conducted via keyword queries ('*coronavirus / Covid / Covid-19 + Italia / crisi / Fase 1 / Fase 2*'; *'coronavirus / Covid / Covid-19 + Italy / crisis / Phase 1 / Phase 2'*). Results from the queries were manually checked to exclude possible noise. Such a controlled data retrieval is essential for aims of this study, as we felt that a contained but 'clean' corpus was preferable to a much bigger but 'noisier' one (as advocated by Jakubíček et al., 2020 as one possible "mitigation strategy" to avoid spam and noise in web-crawled corpora). The textual genres in the corpus include editorials, investigations, commentaries, and interviews and exclude shorter texts such as daily bulletins. In fact, the latter textual type has a more immediate communicative function: to inform about daily deaths and new infections, rather than provide an in-depth commentary to the general crisis. The complete list of sources for the corpus is available in Appendix 1.

To ensure representativeness and balance, we included both national-level newspapers (e.g. *La Repubblica)* and magazines (e.g. *Internazionale*), local newspapers (e.g. *Il Corriere dell'Umbria*), and blogs both at the national and local level (e.g. *Wallstreetitalia*, *Infonaples*). The resulting corpus is composed of 511 articles, 300 for Phase 1 (25 February – 26 April) and 211 for Phase 2 (27 April – 2 June) respectively, for a total of 422,747 words. The corpus does not aim to encompass the totality of articles published in the time frame selected, but rather aims to be a sample as representative as possible of the genre analysed, following the directions on building small specialised corpora of Koester (2022).

The corpus was subsequently POS tagged and analysed with *R* (R Core Team, 2020) for computational analysis, and *SketchEngine* (Kilgarriff et al., 2014) was used for all corpus-assisted content analysis.

## 3. Topic Models

To investigate the macro-level thematic structure of the corpus, we use topic modelling (TM), a text mining methodology that allows to discover latent semantic and thematic clusters in a corpus. TM is a family of probabilistic algorithms of word counts that retrieve a set of *topics*, i.e. clusters of words that co-occur across documents following certain probabilistic patterns (Blei, 2012). TM are unsupervised learning methods that use both modelling assumptions (defined by the researcher) and text properties (bootstrapped from the corpus) to estimate general semantic themes (i.e. topics) on the basis of word co-occurrences (Ciotti, 2017; Murakami et al., 2017; Combei and Giannetti, 2020). TM is well-attested for the analysis of journalistic language (Jacobi et al., 2016). Specifically, we will use Structural Topic Modeling (STM, Roberts et al., 2016), a recently implemented TM algorithm that also allows to model topic distribution as a function of document-level covariates. The effect of covariates can be statistically evaluated through a regression-like model.[2] In our case, we model the temporal covariates of week (counting from the beginning of the regional lockdowns) and Phase (1 vs 2).

The analysis was conducted using the *stm* R package (Roberts et al., 2016). Metadata used for the corpus documents include phase (1 or 2) and date of publication (phase-week-month-day). For instance, an article published on 27 February 2020 would be coded as "phase1_week1_february_27". To distinguish articles published on the same day, letters are used. Therefore, the second article retrieved for 27 February would be coded as "phase1_week1_february_27b".

Pre-processing of the text removed punctuation, numbers and *hapax* (i.e. words that appear only once in the entire corpus). For the removal of stopwords, an ad-hoc list was created,[3] which also includes the words 'coronavirus' and 'Covid', as they are not informative for our specialised corpus. The stemming option was excluded, as poor performance of stemming techniques – i.e. reducing inflected words to their root – for inflectional languages has been demonstrated in the literature (Singh and Gupta, 2016). The resulting corpus has a lexicon of 13118 types and 109348 tokens.

As the number of topics to be retrieved (K) is highly data and task specific (Chang et al., 2009), we selected a K of 3 as the optimal balance between mathematical fit and data interpretability. A preliminary quantitative selection using the *searchK( )* function was performed over a range of values. K= 3 was then chosen for its near-optimal mathematical fit and ease of interpretation. The topics were then given interpretable labels that reflect their general content (Fig. 1):

---

[2] For a more detailed description of STM characteristics, see Roberts et al., 2016.
[3] Based on the one by Gene Diaz and Arthit Suriyawongkul, available at https://github.com/stopwords-iso/stopwords-it.

1. HEALTH[4]

2. SOCIETY[5]

3. ECONOMY[6]

Although κ=3 was chosen as the optimal value, some overlaps of words across topics are to be expected. This is predominantly due to the inter-relatedness of the topics in the data. This however is quite common in topic models, and does not render the distinction across topics meaningless, as the criterion of exclusivity is still met. That is, high-probability words for each topic tend to either not appear in other topics or to appear as low-probability words (Roberts et al., 2014).

Preliminary exploration revealed that all three topics appear in similar proportions in the data, with a slight prevalence of Economy (topic 3, see Fig. 2).

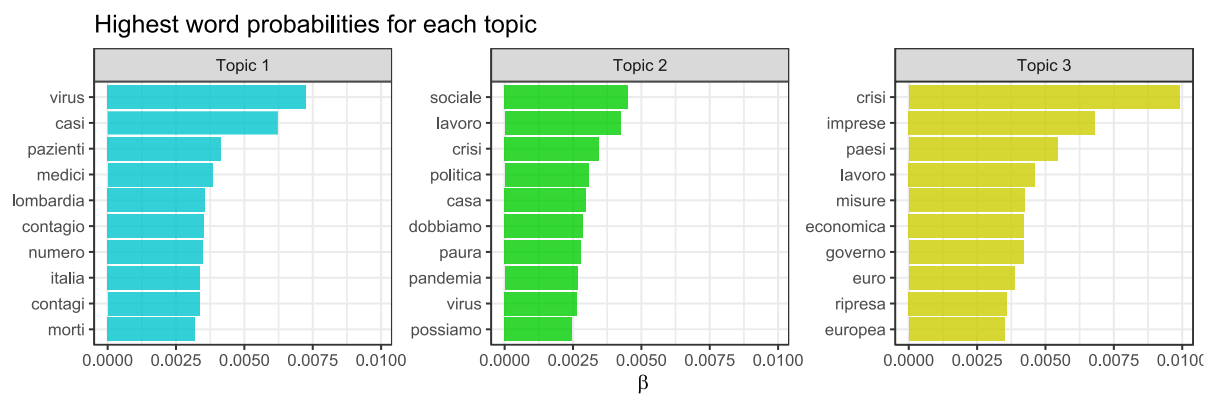

**Figure 1.** Highest word probabilities for each topic

---

[4] Words in the Health topic in Figure 1, in descending order: 'virus, cases, patients, doctors, Lombardy, contagion, number, Italy, contagions, deaths'.
[5] Words in the Society topic in Figure 1, in descending order: 'social, job, crisis, politics, home, we must, fear, pandemic, virus, we can'.
[6] Words in the Economy topic in Figure 1, in descending order: 'crisis, businesses, countries, job, measures, economic, government, euro, growth, European'.

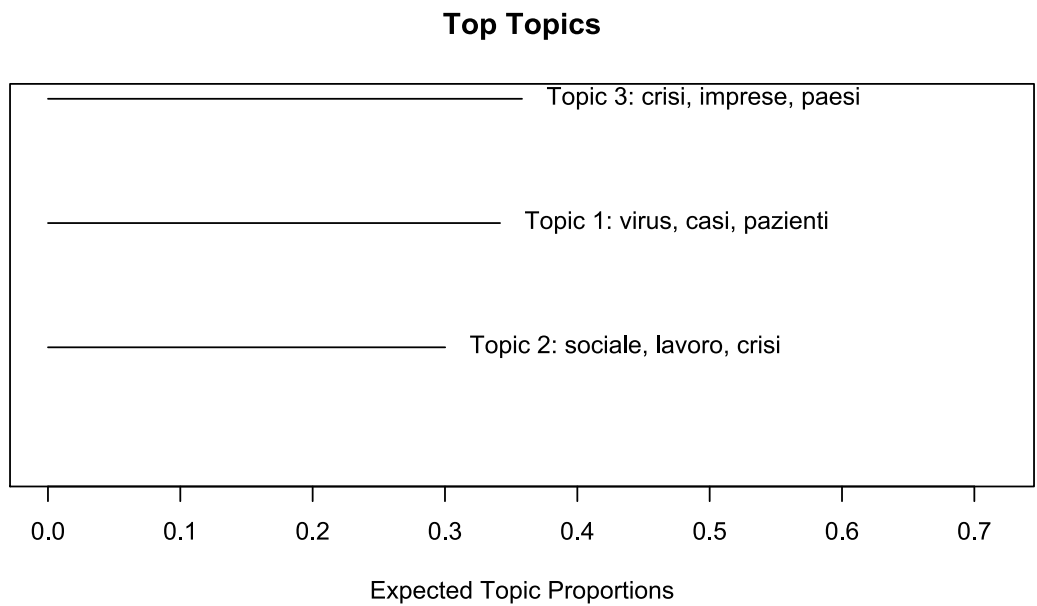

**Figure 2.** Proportions of topics in the corpus

Statistical analysis for topical prevalence (i.e. the frequency with which a specific topic is discussed) was performed on the covariates of Phase and week. We combine the two temporal variables to get a nuanced understanding of how each topic changed across the two phases. Results suggest that a significant shift in discourse happened just before the official beginning of Phase 2: as Health-centred discourse decreases ($p < .0001$), Economy and Society become more prevalent ($p = .05$) (see Fig. 3). Thus, from a discourse centred around the health crisis in Phase 1, Phase 2 witnesses a shift to a discussion more focused on the social and economic repercussions of the pandemic.

By using week as a covariate, we are also able to more precisely locate the turning point. Particularly, Figure 4 shows an inversion of trend, with a spike in prevalence for Economy and a rapid decrease of the topic of Health ($p < .0001$) around week 8 (13-19 April). Societal themes gained more prominence in week 10 (27 April–3 May), where we also see a slight reprise of the topic of Health (however the topic is still significantly less discussed than in week 1, the intercept).

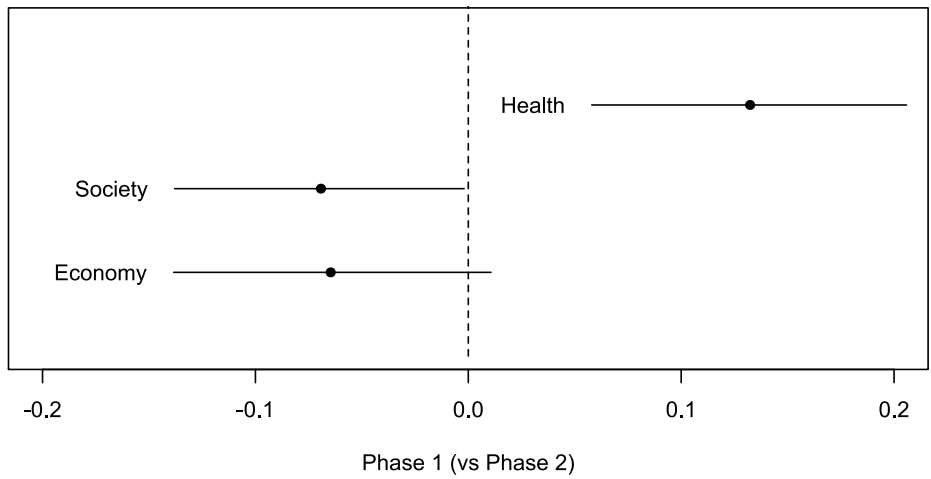

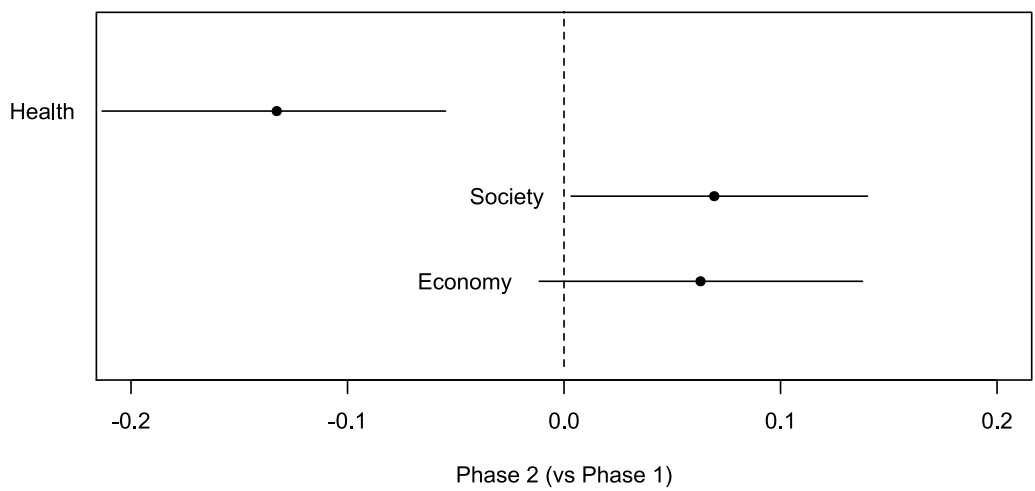

**Figure 3. 3A:** Comparison of topical prevalence in Phase 1 (as compared to Phase 2) and **3B:** in Phase 2 (as compared to Phase 1)

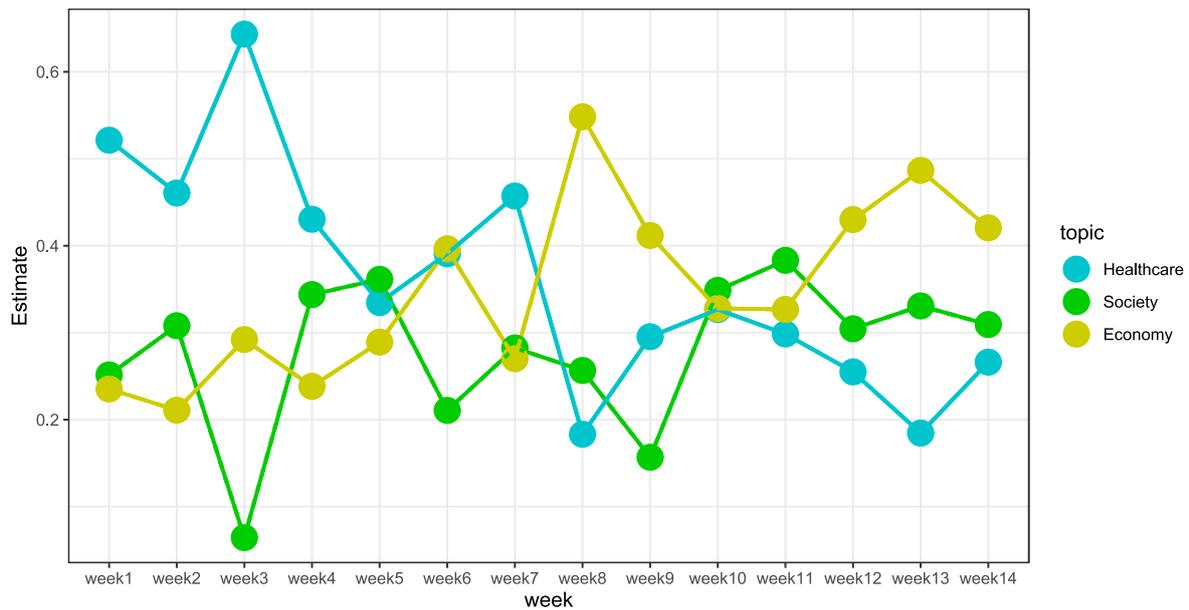

**Figure 4.** Estimates per week of topic proportions

In sum, as the health crisis subsided – with fewer deaths and contagions – mediatic attention shifted toward socio-economic themes. This shift is evident as early as week 5 (23-29 March), but reaches statistical significance in mid-April, around ten days before the announcement of Phase 2 reopening. In the next section we turn to discussing topic-specific metaphors that emerge in our corpus.

**4. Topic-specific metaphors**

We use Conceptual Metaphor Theory (CMT, Lakoff and Johnson, 1980; Semino, 2008; Kövecses, 2010) as a reference framework to investigate figurative language that emerges from each of the three topics retrieved. CMT assumes that metaphorical language is pervasive in human language, and is a cognitive tool through which we interpret and define reality, as "our conceptual system (...) is fundamentally metaphorical in nature" (Lakoff and Johnson, 1980, p. 3). Metaphors are understood to be articulated in three parts: *source domain* – the origin of the metaphorical image (LOVE IS A *JOURNEY*); *target domain* – which the metaphor conceptualises (*LOVE* IS A JOURNEY); and *mapping* – the set of correspondences between the two domains (Kövecses, 2010).

To retrieve possible metaphors in our corpus, we used the representative keywords[7] of our topics '*virus*', '*economia*' ('economy'), '*società*' ('society') and we performed concordance and collocational queries. The queries were manually explored by both authors in order to identify possible topic-specific figurative language. For example, the concordance *'Servirà del tempo per guarire le cicatrici economiche del virus'* ('It will take time to heal the economic scars of the virus') seems to instantiate a conceptual metaphor that equates the economy to a living being, capable of being hurt and scarred. Using the Thesaurus function of Sketchengine, we find synonyms of the metaphorical lexical item identified (in this case 'scar', which finds words such as 'wound', 'hibernation', 'tissue') and explore the distribution of the metaphor in the corpus.

Once a possible metaphor was identified in the text, we turned to the online repository *MetaNet* (MN, Petruck 2018) to check if a conventional metaphor similar to the one identified is present. For the example above, we find the metaphor ECONOMY IS A BODY.[8] More specifically, in our corpus the Economic body is always sick, injured. We decided hence to use the metaphor ECONOMY IS A PATIENT. We also consult the literature to see if this metaphor has been researched to find further examples. If the metaphor that we identify in our corpus is not present in MetaNet, we categorise it as novel.

To ensure that the authors' coding of what constitutes figurative language is consistent and generalisable, a sample of 25 concordances (5 for each macro metaphorical network retrieved, see Table 1 below) and 25 distractors (i.e. concordances taken from the corpus with no conceptual metaphors) was given to an external coder. The coder, a linguistics student and English native speaker, was given the task of coding the concordances as figurative or non figurative in nature. Following Rau and Shih (2021) we performed percentage agreement to test for reliability, as the variables are not independent from each other. We obtained a percentage agreement of 97% between the authors' coding and the external coder, validating the inductive method used by the authors.

---

[7] As plenty of scholars have pointed out in the last decades (see for instance Stubbs, 1996 and Bondi, 2010), the ideological implications of a key lexical element can surface by investigating its collocations and the "semantic preference", namely the tendency of a word to co-occur with other words belonging to a specific semantic category.
[8] https://metaphor.icsi.berkeley.edu/pub/en/index.php/Metaphor:ECONOMY_IS_A_BODY.

We selected all the metaphors emerging from the corpus queries, without deciding *a priori* on a particular set. However, we decided to exclude metaphorical images relating to the semantic domain of WAR, as they have already been thoroughly investigated in other recent studies on the coronavirus pandemic (see Introduction). The imagery of VIRUS IS A WAR is also explored in detail in the "#Reframecovid" Project (Olza et al., 2021; Semino, 2021). We instead focus our attention on several less-investigated metaphorical networks that emerge through the corpus analysis.

After extracting conventional and novel metaphors from our corpus, we further answer RQ2 by postulating the hierarchical metaphorical networks that emerge from the data. Appendix B displays these hypothetical hierarchies.

Table 1 summarises the general metaphors in our corpus.

**Table 1.** Metaphors retrieved in the corpus

| Topic | Metaphors |
|---|---|
| HEALTH (topic 1) | *Target domain*: HEALTH CRISES ARE NATURAL DISASTERS *Source domain*: ECONOMY AND SOCIETY ARE (SICK) BODIES → PATIENTS |

| ECONOMY and SOCIETY (topics 2, 3) | socio-economic CRISES ARE NATURAL DISASTERS; ECONOMY AND SOCIETY ARE BUILDINGS; ECONOMY AND SOCIETY ARE MACHINES; ECONOMY AND SOCIETY ARE (SICK) BODIES → PATIENTS; |
|---|---|

A first noticeable feature of the metaphorical networks retrieved is the overlap across topics. The topic of Health is used both as target domain and as source domain for a socio-economic metaphor (see below Section 5). As target domain, Health is particularly used with the source domain of natural disasters, which is remarkably shared also with Society and Economy metaphors. The use of natural disaster as a metaphor for health crises is well attested for a series of recent health emergencies, such as H1N1, avian flu, SARS, and Ebola (see Introduction). The use of this metaphor for economic crises is also not entirely new, as reported for instance by Cardini (2014). In our corpus, the lemma 'tsunami' – one of the most representative linguistic cues of the metaphor, and the most frequent – is evenly distributed in the corpus (Phase 1: 81 hits, 348.34 pmw; Phase 2: 83 hits, 436.34 pmw). This points to the source domain of natural disaster being consistently used throughout the investigated time span. Moreover, collocational analysis shows that the word 'tsunami' is used as a health metaphor in Phase 1 and as a socio-economic one in Phase 2.[9] The modifiers more frequently recurring with tsunami are in fact '*continuo*' ('continuous', 11 hits) and '*epidemico*' ('epidemic', 2 hits) in Phase 1, and '*economico*' ('economic', 6 hits), '*finanziario*' ('financial', 2 hits) in Phase 2. Moreover, the collocation "X is a tsunami" is used in Phase 1 with the word '*coronavirus*' (2 hits), and in Phase 2 with the word '*crollo*' ('collapse', 1 hit). Furthermore, it is worth noting that the topic of Health takes on the dual role of source and target domain for different metaphors, linking the three topics with shared metaphorical images. Particularly, the metaphor ECONOMY/SOCIETY IS A PATIENT is used pervasively in both Phases, although it is slightly more present in Phase 1. In both phases, journalists talk about economy and society as wounded (*le ferite inflitte all'economia, le piaghe sociali, le ferite al modello sociale:* 'wounds inflicted on the economy', 'the social sores', 'wounds on the social model'). Society and economy are variously characterised by features of living organisms, like stress, fragility, being suffocated (*asfissia*), being 'struck to the heart' (*l'economia colpita al cuore*), having adaptive capacities, being paralysed, etc. The metaphorical mapping of economy and society to (sick) bodies can be seen as a sub-mapping of a more general metaphor ECONOMY AND SOCIETY ARE COMPLEX SYSTEMS, that also includes ECONOMY AND SOCIETY ARE BUILDINGS and MACHINES (see Section 5).

---

[9] Conducted via the Word Sketch Difference tool on SketchEngine. For a more in-depth discussion on the usefulness of this method and for further analyses on the same data, see Busso and Tordini (2021).

Since no recent work on the language of Covid-19 – to the best of the authors' knowledge – focuses on the figurative language used to describe the socio-economic dimension of the pandemic, in the next section we present an in-depth analysis of these topic-specific metaphors.

## 5. Metaphors and Complex Systems: Economy and Society

Seminal works on metaphors by Kövecses (2003, 2010) have already underlined how economic structures and social groups can be thought of as complex abstract systems. More specifically, "a complex abstract system [is conceived] as a nonphysical domain which has many parts that interact with each other in complex ways" (Kövecses, 2010, p. 82). The author also observed that target domains as ECONOMY and SOCIETY are frequently associated with source domains defined as Complex Physical Objects (Kövecses, 2003, p. 98). In particular, metaphors related to complex systems are composed of different submappings, which can be expressed through two conventional conceptual metaphors: COMPLEX SYSTEMS ARE STRUCTURED OBJECTS and COMPLEX SYSTEMS ARE LIVING

ORGANISMS (Kövecses, 2003, p. 106).

The following paragraphs offer a corpus-assisted analysis of the nuances of the representation of the topics/ target domains of Economy (topic 1) and Society (topic 2). We focus specifically on these aspects as – as mentioned above – the Covid crisis has been mainly analysed in its health-related aspects, but we believe that economic and social aspects are crucial to understand the rhetoric around the pandemic. As we will show in the following sections, the conceptual metaphor ECONOMY AND SOCIETY ARE STRUCTURED OBJECTS encompasses two sub-levels, respectively pertaining to buildings/architectural structures (ECONOMY AND SOCIETY ARE BUILDINGS) and machines/devices (ECONOMY AND SOCIETY ARE MACHINES). In this work, the conceptual metaphor ECONOMY AND SOCIETY ARE PATIENTS will be included within the "broader" metaphor ECONOMY AND SOCIETY LIVING ORGANISMS, as motivated in Section 5.3.

*5.1 ECONOMY AND SOCIETY ARE BUILDINGS*

As shown with the STM analysis (section 3), Economy and Society are two central thematic nodes of the journalistic discourse in the first months of the pandemic. This is not surprising, as the strict lockdown measures Italy introduced were constantly discussed in terms of their catastrophic effect on the economy and on the very essence of society and communities. In this context, it seemed appropriate to make some comparisons with the representation of the same target domains that emerged in the main linguistic surveys on the 2008 socio-economic crisis (for a thorough literature review see Arrese and Vara-Miguel, 2016). Indeed, we find many similarities in metaphor use across the two crises, as both the 2008 economic recession and the Covid-19 pandemic have been represented as catastrophic on all levels of contemporary society. Metaphors related to architecture and buildings[10] have proved effective in describing the socio-economic crisis of 2008 (Orts and Rojo, 2009; Rojo and Orts, 2010; Esager, 2011; Joris et al., 2014, 2015, 2018; Arrese and Vara-Miguel, 2016). Through their concreteness, lexical metaphors that map socio-economic structures to buildings show the need for solid foundations (McCloskey, 1995); moreover, the BUILDING source domain also possesses a positive and hopeful connotation of 'progressing toward a goal, toward completion' (Zeng et al., 2021). The health crisis triggered by the Coronavirus has definitively unsettled the foundations of the world's massive capitalist structures. A simple concordance search in our corpus finds that all of the occurrences of *costruire*, 'build' occurs 63 times (149 pmw), '*ricostruire*', 'rebuild' 91 times (215.3 pmw), *distruggere* 'destroy' 26 times (61.5 pmw) are figurative in nature. Framing the Covid crisis as a crisis of the 'foundations' of society and economy effectively represents the threat to the equilibrium of our complex contemporary system caused by the pandemic. The only way out is to start from scratch: '*che si parta dalla BASE. Una casa che crolla non si aggiusta partendo dal tetto*' ('let us start from the FOUNDATION. We cannot rebuild a collapsing house starting from the roof', Exibart 2 April). We are all aware that the building/economy is collapsing. The architect (State, banks, etc.) is now in charge to decide how to reinforce the pillars – for instance, the production process – that support the economic building (Richardt, 2003, p. 275): '*Per l'economia (…) ricostruire significa proteggere le imprese*' ('For the economy (...) rebuilding means protect businesses, Agi 16 April). For several reasons of different natures, the post-Covid reconstruction is felt as extremely complex. Particularly because of the absence of physical ruins: 'as there is no collapsed bridge, road or house, is it necessary to understand what society is to be rebuilt.' ('*È perché non ci sono ponti, strade e case distrutte che occorre capire quale società edificare*', il

---

[10] For a more in-depth discussion around the SOCIETY IS A BUILDING metaphor, see also Goatly (1997), Charteris-Black (2004) and Lu and Ahrens (2008).

Corriere, 7 April). '*Inizia la stagione della ricostruzione, termine improprio visto che il virus che ha mietuto migliaia e migliaia di vittime non ha distrutto nulla di materiale.*' (The season of reconstruction begins, an improper term since the virus that claimed thousands of victims has not destroyed anything material', Avvenire, May 8). This shows an interesting "clash" between literal and figurative. The writers are aware that a metaphorical rebuilding posits many more difficulties than a physical one: there are no actual buildings to reconstruct, but a whole economic and societal system that needs to be somehow restarted.

Awareness of this difficulty inevitably results in a diffused disillusionment: 'we do not rebuild, reimagine, or lay foundations again. We just try to relaunch old devices' (*'non si ricostruisce, non si reimmagina, non si rifonda. Si prova solo a riavviare i vecchi congegni'*, Internazionale, 29 April). This example is particularly useful, as it shows – in line with Kövecses (2003, 2010) – that the conceptual metaphor ECONOMY AND SOCIETY ARE STRUCTURED OBJECTS not only includes lexical metaphors relating to buildings or architectural structures, but also to machines or devices. This specific aspect will be further explored in the following paragraph.

*5.2 ECONOMY AND SOCIETY ARE MACHINES*

Several international studies on the 2008 economic crisis have also shown a particularly high occurrence of the conceptual metaphor ECONOMY IS A MACHINE (see Orts and Rojo 2009; Rojo and Orts 2010; Esager 2011; Horner 2011; Joris et al. 2015; Joris et al. 2018). These works demonstrate through a quantitative and/or qualitative approach that economy can be also conceived as a machine or a device consisting of several components. Each gear is fundamental to the functioning of the complex system. In crisis situations, however, the mechanical object suffers serious failures which compromise its functioning or movement. In this respect, Kövecses (2003: 98) notes that THE LASTINGNESS OF THE COMPLEX SYSTEM IS THE STRENGTH OF THE OBJECT; THE FUNCTIONING OF THE COMPLEX SYSTEM IS THE FUNCTIONING OF THE OBJECT.

The numerous examples present in our corpus are indicative of the effectiveness of this metaphor. Money is at times defined as 'the fuel that keeps the globalised world on its feet' (*'il carburante che tiene in piedi il mondo globalizzato'*, La Repubblica, 14 March). However, the engine is the most important component, as its functioning is essential for the functioning of the socio-economic machine. In the mediatic discourse on Covid-19 in Italy, it has been identified with:

– Enterprises, for instance, 'the bureaucratic disease caused by Covid-19 stops the enterprises, the engine of recovery' ('*la burodemia (malattia da burocrazia) da Covid-19 ferma le imprese, motore della ricerca*', Confartigianato.it, 28 April);

– Eco-sustainable energy, namely the 'green engine of recovery' ('*verde motore della ripresa*', europa.today, 21 May). In some cases, Covid-19 is thought as an opportunity to replace the old engine powered by fossil fuels with a new green engine;

– Women, who can be the 'the engine of Phase 2' ('*Le donne possono essere il motore della Fase 2*', Fortune Italia, 22 April). This choice is due to internal reasons related to the health crisis, as women showed lower infection rates.

In sum, we can say that the restarting of the blocked/damaged system can be guaranteed only if the engine is repaired or replaced.[11] Otherwise, the reopening of economic and social activities would be jeopardised.

*5.3 ECONOMY AND SOCIETY ARE LIVING ORGANISMS*

As already noted in Section 4, economy and society, which have been struck by a cataclysm, are also conceived as sick and/or wounded organisms – that is, as patients who need immediate care.[12] This personification process allows to concretise the disease and hence to declare the urgency of a therapy. Similarly to what has been reported in some studies on the 2008 crisis (see for instance Wang et al., 2013; Cardini, 2014)[13], in the scenario of the Covid-19 crisis State and banks frequently become health personnel treating sick corporations. Exponents of the *Ordine dei Commercialisti ed Esperti Contabili*[14] report that while the 'doctors' (government, banks and State) are researching the cure, 'patients' (Italian SMEs) are at risk of dying due to the bureaucratic complexity of banking procedures (Il Roma, 25 April). At times, we also observe thought-provoking co-occurrences of different conceptual metaphors, often pertaining to both natural cataclysms and organisms, for instance: '*lo tsunami è stato superato. Ne rimangono i segni e le cicatrici, ma il peggio è passato*' (Open, 28 April) ('the tsunami is overcome. The marks and scars remain, but the worst is over'); '*questa tempesta partorirà una società paritetica*' (Collettiva, 28 March) ('this tempest will give birth to a more equal society'). In both passages, the two metaphors of natural disasters and bodies are coalesced to create a syncretic and more powerful image.

---

[11] As already claimed by Lu and Ahrens (2008, p. 394) the system "can be conceived either as an old building that needs to be reconstructed, or as a new building that needs to be built".

One of the most wide-spread instances of this metaphor is the use of the words '*ferita*' ('wound') and '*cicatrice*' ('scar') to describe the effects of the pandemic on both economy and society: '*gli infermieri sono il cuore della nostra comunità ferita*' (Il Messaggero, 25 March), '*oggi s'è vista l'Italia ferita*' (Huffington Post, 18 May) ('nurses are the heart of our wounded community', 'today we saw a wounded Italy').

Interestingly, not only economy and society are described as patients and wounded bodies, but often the metaphors draws from symptoms of Covid infections: '*il paese moribondo riprenda*

---

[12] In literature, the ECONOMY IS A SICK PERSON metaphor is defined as "a metaphor whose source domain is the functioning of the human body which is mapped on the society's economic life" (Silaški and Đurović, 2010, p. 133).
[13] Wang et al. (2013) noticed that in British and Russian economic newspapers the economy is frequently represented as a patient, whereas the doctor – that is, the State or the economists – has the task of stopping the spreading of the disease. Cardini (2014) also showed that some English newspapers conceptualise the economy as an organism exposed to serious health risks, severely debilitated, or on the verge of death.
[14] Order of Accountants and Accounting Experts.

*a respirare velocemente – prima che i danni siano permanenti*' (Trieste News, 1 April); '*Miliardi di soldi pubblici che sono l'ossigeno da somministrare in grandi dosi (...) altrimenti il corpo socio-economico non respira*' (Calciomercato.it, 2 May); '*La stessa definizione scelta, distanziamento "sociale" invece che "fisico", può, forse, funzionare come antidoto virale* (La Repubblica, 25 May), ('the dying country must start to breathe again quickly - before the damage is permanent'; 'billions of public money that are the oxygen to be administered in large doses (...) otherwise the socio-economic body cannot breathe'; 'the same definition chosen - '"social" rather than "physical" distancing – can, perhaps, work as a viral antidote').

A compelling picture of the situation is provided by the socio-political journal Jacobin (3 April), which underlines 'the need to keep economy in a pharmacological coma until the pandemic is over' (*'necessità di tenere in coma farmacologico l'economia in attesa di uscire dalla crisi sanitaria'*). Paraphrasing Tooze, the authors convey a peculiar image: 'the economy is to be treated like patients in intensive care, who must be kept alive with minimal vital functions for a few months' – i.e., the waiting time for a cure ('*bisogna fare con l'economia ciò che si fa con i pazienti in terapia intensiva, bisogna tenerli in vita con funzioni vitali minime per qualche mese*').[15]

Besides the image of the patient in a coma, the suspension of vital functions can take the form of hibernation: '*le misure di contenimento messe in campo dai Paesi membri (...) hanno messo l'economia in uno stato di ibernazione*' (Euronews, 6 May); '*non è così che si riuscirà risvegliare la società dal letargo in cui il coronavirus l'ha precipitata*' (Start Magazine, 1 April) ('the containment measures established by the EU member states have put the economy in hibernation'; 'this is not how society will be able to awaken from the hibernation into which coronavirus has plunged it').

Overall, the general conceptual metaphor of ECONOMY AND SOCIETY ARE LIVING

---

[15] The full article is available in Italian at https://jacobinitalia.it/solo-la-politica-puo-evitare-lapocalisse/.

ORGANISMS is used to describe the poor state of health of socio-economic bodies. Different submappings are used to highlight different aspects of the crisis and some proposed solutions. The use of health-related metaphors is particularly relevant in the case of the Covid-19 crisis, in which physical effects of the viral infection are used as figurative representations of the damages to societies and economies.

In the following paragraph, we will integrate the qualitative observations on figurative language in Italian newspapers with the results of a corpus-assisted content analysis.

## 6. Content analysis

For a more in-depth and data driven look at the analysed metaphors, we further explore collocational patterns of the two keywords 'economia' and 'società' using the *WordSketch* tool of the online corpus software *SketchEngine* (Kilgarriff et al., 2014). Collocational analysis is a widespread tool in corpus-assisted discourse analysis, as it reveals patterns of word usage that are salient to the topic under investigation (Baker, 2006; Baker et al., 2013). Furthermore, usage patterns can also show attitudes and evaluations towards the node (i.e. the investigated word) (Baker, 2016; Stubbs, 2001). Analysing the lexical items that co-occur with our topic-specific keywords can hence enhance our understanding of their representation in the corpus (Baker, 2006, p. 96).

Using the WordSketch tool, we obtained a graphical display of the collocations, based on their grammatical relationships – with a threshold of 10 collocates for each type of relationship.[16] We considered three types of collocational relations: (1) collocates that are modifiers of the node and (2) verbs that have the node as a subject or (3) as an object. After calculating statistical association for each collocate, the software assigned a (log)Dice score[17] to each association:

---

[16] A threshold of 9 is used to avoid overpopulated graphs. However, we also include in our analysis examples not shown in Figures 5 and 6.

[17] In SketchEngine, the Dice Score is expressed through a logarithmic scale: "theoretical maximum is 14, in case when all occurrences of *X* co-occur with *Y* and all occurrences of *Y* co-occur with *X*. Usually the value is less than

the greater the strength of association of the two words in the corpus, the greater the size of the bubble in the graph, and the closer to the centre.

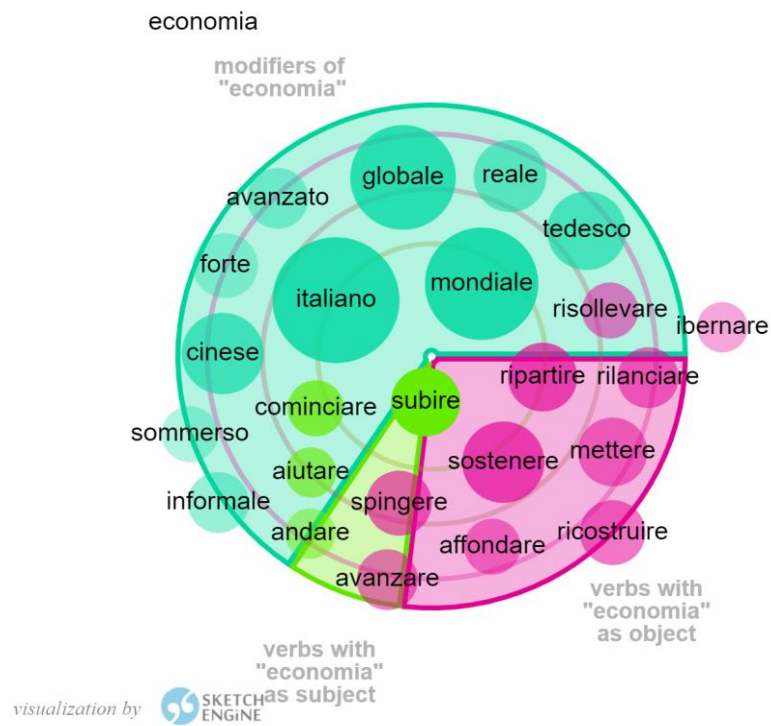

**Figure 5.** Wordsketch of the word '*economia*' ('economy')

---

10." (Rychlý, 2008, p. 9). Therefore, a lower score indicates that the words in the collocation also frequently occur with other words in the corpus.

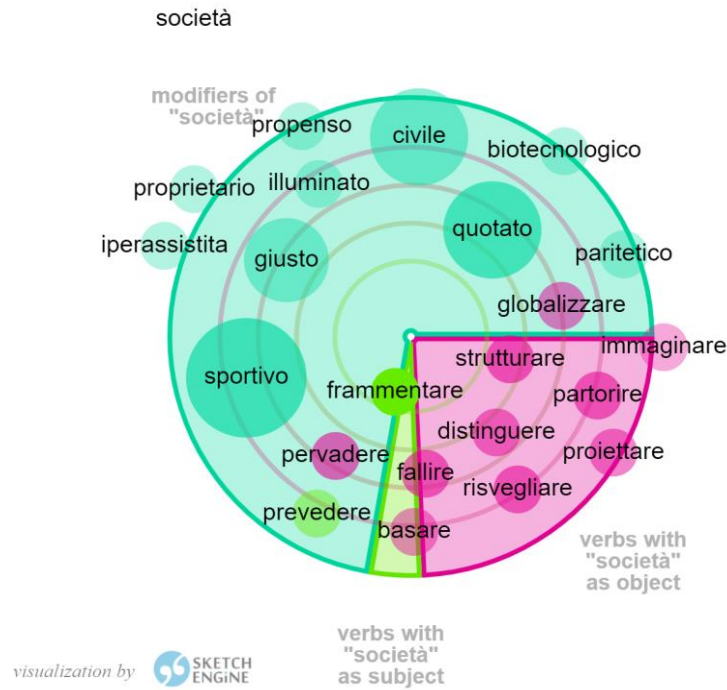

**Figure 6.** Wordsketch of the word '*società*' ('society')

In Figure 5 we can see the word sketch of *'economia'*. As for modifiers (light blue section), 'economy' is mainly connoted by geo-political adjectives ('*italiano*', 'Italian' – logDice 11.17; '*globale*', 'global' – logDice 10.95; '*cinese*', 'Chinese'- logDice 10.39). Slightly lower scores – but still noteworthy – are assigned to modifiers that reflect the metaphorical mappings above analysed: *'sommerso'* ('submerged' – logDice 8.93) which pertains to the 'natural disaster' source domain, 'forte' ('strong'- logDice 9.1), and (not shown) *'debole'* ('weak' – logDice 8.85) and '*malato*' ('ill' – logDice 8.8) that refer to the 'living organism' source domain. Natural disasters and (sick) bodies appear then to be the two source domains which are mostly referred to in adjectival modification.

On the other hand, the semantic scope of modifiers for società (Figure 6) is more wide-ranging. In fact, 'society' is understood not only as a group/organisation of individuals ('*civile*', 'civil' – logDice 10.2), but also as an association ('*sportivo*', 'sporty' – logDice 11.34) and as an economic activity ('*quotato*', 'listed' – logDice 11.22; '*capitalistico*' 'capitalist' – logDice 9.24). Interestingly, no adjective within the first 10 collocates is figurative in nature – differently from *'economia'* in Figure 5. Among the verbs with economy as subject (green section), we notice in Figure 5 the verb '*subire*' ('to suffer', 'to undergo') (logDice 11.91). In this respect, it is to note that a collocational analysis recently performed by Papapicco (2020) on the keyword 'Coronavirus' already suggested a recurrent representation of passivity in Italian newspapers. In this case, the agent of the action (economy) is a patient who is not strong enough to react to the damages caused by external entities. Society is interestingly used as subject with the verb '*frammentar(si)*' ('to shatter' – logDice 13.4). This verb evokes the metaphor of society as a structure on the verge of breaking.

Finally, verbal collocates with economy and society as objects (pink section in Figures 5 and 6) are the area in which figurative language predominantly emerges. The majority of collocates in fact relates to the conceptual metaphors ECONOMY AND SOCIETY ARE STRUCTURED OBJECTS and ECONOMY AND SOCIETY ARE LIVING ORGANISMS. This is in line with our preliminary observations during data collection (see section 4).

Within the submapping ECONOMY AND SOCIETY ARE BUILDINGS, which belongs to the first metaphorical domain, we find verbs such as '*sostenere*' ('to support' – logDice 10.57), '*ricostruire*' ('to rebuild' – logDice 9.93), '*affondare*' ('to sink', 'to plunge' – logDice 9.7) for Economy, and '*strutturare*' ('to structure' – logDice 11), '*basare*' ('to base' – logDice 10), '*frammentare*' ('to fragment' – logDice 9.89) for Society. Verbs belonging to the submapping ECONOMY AND SOCIETY ARE MACHINES are also used for the economy, particularly '*ripartire*' ('to restart' – logDice 10.47) and '*spingere*' ('to push' – logDice 10.18). Economy is hence often represented in the data as a jammed vehicle that needs a push to regain functionality and movement. This is in line with previous studies of economic crises, especially the 2008 crisis (see Section 5.2), that found how metaphors of structured objects are particularly useful to depict both the collapse of the capitalist structure and the need for the financial market to start moving again. For the conceptual metaphor ECONOMY AND SOCIETY ARE LIVING ORGANISMS, we see the verb '*ibernare*' ('to hibernate' – logDice 9.15) already mentioned in Section 5.3. A lower – yet still notable – logDice score (8.9) was found for the verb '*infettare*' ('to infect'). On the other hand, we can see that society is used with verbs such as '*partorire*' ('to give birth' – logDice 10.91),'*risvegliare*' ('awaken' – logDice 10.9) and '*generare*' ('to generate' – logDice 9.2) These results seem to point to a difference in patterns of metaphorical representations of economy and society: while during crises economy is often represented as an injured organism, a sick patient that needs tending to (see Cardini 2014), society is portrayed as an unborn or sleeping organism, in need of a rebirth.

Overall, collocational analysis shows that metaphors are mostly expressed syntactically by verbs that take economy and society as direct objects. This finding is in line with previous literature that found that predicate-argument constructions are the most common grammatical means of communicating a metaphor (Sullivan 2009, 2016).

## 7. Conclusions

We presented a study on online media language on the Covid-19 pandemic, in which we contrasted topics and metaphors used by the press in the first two phases of the Italian response to the pandemic. The current study is innovative in several respects: we focused on Italian, a less-researched language and the language of the first European country to be hit by the pandemic in February 2020; we looked at the Covid pandemic in its entirety, encompassing not only the sanitary emergency but also the concurrent social and economic crises; finally, we combined metaphor analysis with the computational technique of Structural Topic Modelling – which results in an unusual yet effective combination of qualitative and quantitative methodologies.

Notably, our STM analysis retrieved three main general topics: Health, Economy, and Society, consistently with the analysis of US social media and news outlets in March 2020 presented in Chipidza et al. (2022). Further statistical analyses revealed that there are significant differences in topic prevalence between the two phases, and that the shift in discourse antedates the official beginning of Phase 2 by roughly two weeks.

After having identified the macro themes in the data, we focused on topic-specific metaphors, i.e. metaphors that use the analysed topics as target domains. The analysis of topic-specific metaphors allows us to see how figurative language has been used in Italian press and which aspects of the Covid pandemic were preferred in journalistic discourse. Notably, we found a substantial overlap in metaphor use: Health is both used as a source and target domain, and the source domain of 'natural disaster' is employed for both Health and socio-economic metaphors. We argue that this overlap and reuse of the same source and target domains for different topic-specific metaphors is functional to the creation of a shared "metaphorical lexicon". This would allow journalists to anchor new concepts and themes to figurative representations already familiar to the reader. Some metaphorical phrases – such as the "tsunami" effect of the virus and of the economic crisis, the "wounds" and "scars" left by the virus, the necessity for a "restart" of the economic engine – become almost formulaic and fixed in nature. In other words, repetition and reuse of certain metaphors may help in shaping the public's attitude toward the ongoing crisis of the pandemic (Krennmayr, 2015).

Given that most of the literature's attention has been focused on the Covid crisis as a health crisis – somehow disregarding the social and economic aspects of lockdowns and regulations – we further presented a more in-depth analysis of socio-economic metaphors building on the paradigm proposed by Kövecses (2003, 2010). This framework holds that both economic and social structures can be conceived as complex systems, in the form of STRUCTURED OBJECTS and LIVING ORGANISMS. This paradigm perfectly aligns with our inductive results from the corpus analysis (see Table 1). Particularly, we find in our corpus different submappings of these two more abstract metaphors: Society and Economy are often conceptualised as either architectural structures, machines/devices, and (sick) living bodies. These metaphors are very well attested in the literature on economic crises such as the 2008 recession, but they have never been analysed – to the best of the authors' knowledge – in the context of a health emergency.

Based on previous studies in the relevant literature and with the help of the online repository MetaNet, we also reconstructed the (possible) metaphorical networks of topic specific metaphors (see Appendix B).

A corpus-assisted content analysis confirmed the pervasiveness of the metaphors of Economy and Society as STRUCTURED OBJECTS and LIVING ORGANISMS. Namely, in our corpus Economy and Society are pictured either as a jammed vehicle, a collapsed building, or as (suffering) living bodies. The use of health as source domain for Economy and Society correlates with the depiction of a static system (see also the use of the verb '*ibernare*', 'to freeze') that needs to be kept in stable conditions. Furthermore, the lexical cues of this metaphor (see Section 5.3) also mirror the clinical consequences of Covid-19. As for lexical cues of Society metaphors, we found verbs (with '*società*' as object) with both a negative meaning – such as '*fallire*' ('to fail') and '*frammentar(si)*' ('to shatter') – and a more positive one – like '*generare*', '*partorire*' ('to generate', 'to give birth'), '*risvegliare*' ('to awaken'), and '*immaginare*' ('to imagine'). Compared to Economy, however, the rhetoric surrounding Society appears overall more optimistic: despite the severe consequences of the health crisis, a rosier future seems to be possible.

In sum, findings from this study revealed how figurative language is used to discuss the different thematic nodes of the Covid-19 pandemic: Health, Society, and Economy. These themes are not equally used in both phases, but their prevalence shifts from a more health-centred discourse in the early pandemic days to a greater attention to economic and societal facets. Modelling the shift in topic prevalence also allowed us to observe the shift in topic-related metaphors, as recent studies have highlighted how metaphor change can reflect societal change (De Landtsheer, 2015; Nerghes et al., 2015; Burgers, 2016; Burgers and Ahrens, 2020; Zeng et al., 2021).

Particularly, we see that online media draws on a well-established inventory of different metaphorical networks, attested for previous health and socio-economic crises, and combines them with a few novel submappings that emerge from the current emergency. We also show how these metaphors are used in innovative ways to adequately conceptualise the current crisis, creating a "formulary" of metaphorical images that blend different source and target domains and create overlaps across different metaphors and different topics. These overlaps both help "bridge" the shift in discourse between the two phases, discussing the societal and economic crisis with health metaphors, and also mirror the close intersection across the different aspects of the Covid pandemic: social, economic, and sanitary. In other words, our analysis shows the complexity of the Covid crisis and of its repercussions on every aspect of society. Our study also clearly shows that analysing longitudinal changes in metaphors in discourse is an effective method to analyse conceptualisations of societal change.

**References**

Angeli, E. L. (2012). Metaphors in the Rhetoric of Pandemic Flu: Electronic Media Coverage of H1N1 and Swine Flu. *Journal of Technical Writing and Communication*, *42(3)*, 203–222.

Arrese, Á. (2015). Euro crisis metaphors in the Spanish Press. *Communication & Society*, *28*(2), 19– 38.

Arrese, Á., & Vara-Miguel, A. (2016). A comparative study of metaphors in press reporting of the Euro crisis. *Discourse & Society*, *27*(2), 133-155.

Bagli, M. (2021). #HeroesWearMasks. In L. Busso and O. Tordini (Eds.), *Describing The New*

*Pandemic: Linguistic And Communicative Studies On The Covid-19 Crises.* In L. Busso and O. Tordini (Eds.), *Describing the New Pandemic: Linguistic And Communicative Studies On The Covid-19 Crises. Rassegna Italiana di Linguistica Applicata, 53* (1), 67–90.

Baker, P. (2006). *Using corpora in discourse analysis*. London: Continuum.

Baker, P. (2016). The shapes of collocation. *International Journal of Corpus Linguistics*, *21*(2), 139-164.

Baker, P., Gabrielatos, C., & McEnery, T. (2013). *Discourse analysis and media attitudes: The representation of Islam in the British press*. Cambridge University Press.

Balteiro, I. (2017). Metaphor in Ebola's popularized scientific discourse. *Iberica, 34*, 209–230.

Baroni, M., & Bernardini, S. (2005). BootCaT: Bootstrapping corpora and terms from the web. In M.T. Lino, M.F. Xavier, F. Ferreira, R. Costa & R. Silva (Eds.), *Proceedings of Fourth International Conference on Language Resources and Evaluation (LREC 2004)* (pp. 1313– 1316). Lisbon: ELDA.

Benzi, M., & Novarese, M. (2022). Metaphors we Lie by: our 'War' against COVID-19. *History and Philosophy of the Life Sciences*, *44*(18), 1–22.

Besomi, D. (2019). The metaphors of crises. *Journal of Cultural Economy*, *12*(5), 361–381.

Blei, D.M. (2012). Probabilistic topic models. *Communications of the ACM*, *55*(4), 77–84.

Bondi, M. (2010). Perspectives on keywords and keyness. In M. Bondi & M. Scott (Eds.), *Keyness in Texts* (pp. 1–20). Amsterdam: John Benjamins.

Burgers, C. (2016). Conceptualizing Change in Communication through Metaphor. *Journal of Communication*, 66(2), 250–265.

Burgers, C., & Arens, K. (2020). Change in metaphorical framing: metaphors of trade in 225 years of state of the union addresses (1790–2014). *Applied Linguistics, 41*(2), 260–279.

Busso, L., & Tordini, O. (2021). "Ricostruire senza macerie". La crisi sanitaria da Covid-19

nel linguaggio giornalistico italiano tra fase 1 e fase 2 ["Rebuilding without debris". The Covid-19 crisis in Italian newspapers between Phase 1 and Phase 2]. In L. Busso & O. Tordini (Eds.), *Describing the New Pandemic: Linguistic And Communicative Studies On The Covid-19 Crises. Rassegna Italiana di Linguistica Applicata, 53* (1), 45-66.

Camus, J.T.W. (2009). Metaphors of cancer in scientific popularization articles in the British press. *Discourse Studies*, *11*(4), 465–495.

Cardini, Filippo-Enrico (2014). Analysing English metaphors of the economic crisis. *Lingue & Linguaggi, 11*, 59–76.

Chang, J., Gerrish, S., Wang, C., Boyd-Graber, J. L., & Blei, D. M. (2009). Reading tea leaves: How humans interpret topic models. In Y. Bengio, D. Schuurmans, J.D. Lafferty, C.K.I. Williams & A. Culotta (Eds.), *Advances in neural information processing systems 22* (pp. 288–296). Cambridge, MA: MIT Press.

Charteris-Black, J. (2004). *Corpus Approaches to Critical Metaphor Analysis*. New York: Macmillan.

Chipidza, W., Akbaripourdibazar, E., Gwanzura, T., &. Gatto, N.M. (2022). Topic Analysis of Traditional and Social Media News Coverage of the Early COVID-19 Pandemic and Implications for Public Health Communication. *Disaster Med Public Health, 16*(5), 1881–1888.

Ciotti, F. (2017). What's in a Topic Model? Critica teorica di un metodo computazionale per l'analisi del testo. [What's in a Topic Model? Theoretical critique of a computational method of text analysis]. *TESTO & SENSO*, *18*, 1–11.

Citron, F.M., & Goldberg, A.E. (2014). Metaphorical sentences are more emotionally engaging than their literal counterparts. *Journal of cognitive neuroscience*, *26*(11), 2585–2595.

Combei, C.R., & Giannetti, D. (2020). The Immigration Issue on Twitter Political Communication. Italy 2018-2019. *Comunicazione politica*, *21*(2), 231–263.

Demjén, Z., & Semino, E. (2016). Using metaphor in healthcare. *The Routledge handbook of metaphor and language*, 385–399.

De Landtsheer, C. (2015). Media rhetoric plays the market. The logic and power of metaphors behind the financial crises since 2006. *Metaphor and the Social World, 5*(2), 205–222.

Esager, M. (2011). Fire and water – A comparative analysis of conceptual metaphors in English and Danish news articles about the credit crisis 2008. Retrieved from http://pure.au.dk/portal/files/40317984/Fire_and_Water.pdf.

Gibbs, R.W. (Ed.) (2008). *The Cambridge Handbook of Metaphor and Thought*, Cambridge: Cambridge University Press.

Goatly, A. (1997). *The language of metaphors*. London/New York: Routledge.

Hanne, M., & Hawken, S.J. (2007). Metaphors for illness in contemporary media. *Medical Humanities, 33*, 93–99.

Hellsten, I. (2000). Dolly: Scientific breakthrough or Frankenstein's monster? Journalistic and scientific metaphors of cloning. *Metaphor and Symbol, 15*, 213–221.

Huang, M., & Holmgreen, L.L. (Eds.). (2020). *The Language of Crisis: Metaphors, frames and discourses*. Amsterdam: John Benjamins.

Horner, J.R. (2011). Clogged systems and toxic assets. News metaphors, neoliberal ideology, and the United States 'Wall Street Bailout' of 2008. *Journal of Language and Politics*, *10*(1), 29–49.

Jacobi, C., van Atteveldt, W., &. Welbers, K. (2016). Quantitative analysis of large amounts of journalistic texts using topic modelling. *Digital Journalism*, *4*(1), 89– 106.

Jakubíček, M., Kovář, V., Rychlý, P., & Suchomel, V. (2020). Current challenges in web corpus building. In Barbaresi, A., Bildhauer, F., Schäfer, R., & Stemle, E, *Proceedings of the 12th Web as Corpus Workshop*. *Proceedings of the 12th Web as Corpus Workshop* (pp. 1– 4). ELRA.

Joris, W., d'Haenens, L., & Van Goro, B. (2014). The euro crisis in metaphors and frames: Focus on the press in the Low Countries. *European Journal of Communication, 29*(5), 608–617.

Joris, W., Puustinen, L., Sobieraj, K., & d'Haenens, L. (2015). The battle for the Euro: Metaphors and frames in the Euro crisis news. In R. G. Picard (Ed.), *The Euro crisis in the media. Journalistic coverage of economic crisis and European Institutions* (pp. 127–150). London: I. B. Tauris & Co.

Joris, W., Puustinen, L., & d'Haenens, L. (2018). More news from the Euro front: How the press has been framing the Euro crisis in five EU countries. *The International Communication Gazette*, 1–19.

Kennedy, V. (2000). Metaphors in the news – introduction. *Metaphor and Symbol* 15, 209–211.

Kilgarriff, A., Baisa, V. Bušta, J., Jakubíček, M., Kovář, V., Michelfeit, J., Rychlý, P., & Suchomel, V. (2014). The Sketch Engine: ten years on. *Lexicography, 1,* 7–36.

Koester, A. (2022). Building small specialised corpora. In A. O'Keeffe & M. McCarthy, (Eds). *The Routledge Handbook of Corpus Linguistics* (pp. 48-61). Routledge.

Kövecses, Z. (2003). *Metaphor and Emotion: Language, Culture, and Body in Human Feeling*. Cambridge: Cambridge University Press.

Kövecses, Z. (2010). *Metaphor: A Practical Introduction*. Oxford: Oxford University Press.
Lakoff, G. J., & Johnson, M. (1980). *Metaphors We Live By*. Chicago: University of Chicago Press. Krennmayr, T. (2015) What Corpus Linguistics Can Tell Us About Metaphor Use In Newspaper Texts. *Journalism Studies*, *16*(4), 530–546.

Larson, B., Nerlich, B., & Wallis, P. (2005). Metaphors and biorisks: The war on infectious diseases and invasive species. *Science communication*, *26*(3), 243–268.

López, J. (2003). *Society and its Metaphors*. New York/London: Continuum.

Lu, L.W., & Ahrens, L. (2008). Ideological influence on BUILDING metaphors in Taiwanese presidential speeches. *Discourse & Society, 19*(3), 383–408.

Maasen, S., & Weingart, P. (1995). Metaphors--Messengers of Meaning: A Contribution to an Evolutionary Sociology of Science. *Science Communication*, 17, 9–31.

Martinez-Brawley, E., & Gualda, E. (2020). Transnational Social Implications of the Use of the "War metaphor" Concerning Coronavirus: A Birds' Eye View. *Culture e Studi del Sociale*, *5*(1), 259–272.

McCloskey, D. (1995). Metaphors Economists Live By. *Social Research, 62*(2), 215–237.

Murakami, A., Thompson, P., Hunston, S., & Vajn, D. (2017). 'What is this corpus about?': using topic modelling to explore a specialised corpus. *Corpora*, *12*(2), 243–277.

Nerlich, B. & Halliday, C. (2007). Avian Flu: The Creation of Expectations in the Interplay between Science and Media. *Sociology of Health & Illness, 29*(1), 46–65.

Nerghes, A., Hellsten, I., & Groenewegen, P. (2015). A toxic crisis: Metaphorizing the financial crisis. *International Journal of Communication*, *9*, 106–132.

Nünning, A. (2012). Making Crises and Catastrophes – How Metaphors and Narratives Shape their Cultural Life. In C. Meiner, & K. Veel (Eds.). *The Cultural Life of Catastrophes and Crises* (pp. 59–88). Amsterdam: de Gruyter.

Olza, I., Koller, V., Ibarretxe-Antuñano, I., Pérez-Sobrino, P., & Semino, E. (2021). The# ReframeCovid initiative: From Twitter to society via metaphor. *Metaphor and the Social World*, *11*(1), 98–120.

Orts, M.Á., & Rojo, A.M. (2009). Metaphor framing in Spanish economic discourse: A corpus-based approach to metaphor analysis in the Global Systemic Crisis. Paper presented at the First International Conference of AELINCO (Corpus Linguistics Spanish Association), University of Murcia, 7-9 May 2009.

Papapicco, C. (2020). Informative Contagion: The Coronavirus (Covid-19) in Italian

journalism. *Online Journal of Communication and Media Technologies, 10*(3), 1–12.

Pelclová, J., & Lu, W. L. (Eds.). (2018). *Persuasion in Public Discourse: Cognitive and functional perspectives* (Vol. 79). Amsterdam: John Benjamins.

Petruck, M. (2018). *MetaNet*. Amsterdam: John Benjamins. Available at: https://metanet.icsi.berkeley.edu/metanet/.

Rau, G., & Shih, Y. S. (2021). Evaluation of Cohen's kappa and other measures of inter-rater agreement for genre analysis and other nominal data. *Journal of English for academic purposes*, 53/101026.

Richardt, S. (2003). Metaphors in expert and common-sense reasoning. In C. Zelinsky-Wibbelt (Ed.), *Text, Context, Concepts* (pp. 243– 296). Berlin: de Gruyter.

Roberts, M. E., Stewart, B. M., Tingley, D., Lucas, C., Leder-Luis, J., Gadarian, S. K., ... & Rand, D. G. (2014). Structural topic models for open-ended survey responses. *American journal of political science*, *58*(4), 1064-1082.

Roberts, M. E., Stewart, Brandon M., & Tingley, D. (2016). Navigating the Local Modes of Big Data: The Case of Topic Models. In R. M. Alvarez (Ed.), *Computational Social Science: Discovery and Prediction (Analytical Methods for Social Research)* (pp. 51– 97). Cambridge: Cambridge University Press.

Rojo, A.M., & Orts, M.Á. (2010). Metaphorical pattern analysis in financial texts: Framing the crisis in positive or negative metaphorical terms. *Journal of Pragmatics, 42*, 3300– 3313.

Rychlý, P. (2008). A Lexicographer-Friendly Association Score. In P. Sojka & A. Horák (Eds.), *Proceedings of Recent Advances in Slavonic Natural Language Processing*, RASLAN 2008, (pp. 6– 9). Masaryk University, Brno.

Sabucedo, J.M., Alzate, M. & Hur, D. (2020). Covid-19 and the metaphor of war. *International Journal of Social Psychology*, *35*(3), 618– 624.

Semino, E. (2008). *Metaphor in discourse*. Cambridge: Cambridge University Press.

Semino, E. (2021). "Not soldiers but fire-fighters"–metaphors and Covid-19. *Health communication*, *36*(1), 50–58.

Silaški, N., & Đurović, T. (2010). The Conceptualisation of the Global Financial Crisis Via the Economy Is A Person Metaphor – A Contrastive Study of English And Serbian. *Facta Universitatis, 8*(2), 129–139.

Singh, J., & Gupta, V. (2016). Text stemming: Approaches, applications, and challenges. *ACM Computing Surveys (CSUR)*, *49*(3), 1–46.

Sontag, S., (1990). *Illness as metaphor*. New York: Anchor.

Stubbs, M. (1996). *Text and Corpus Analysis: Computer-assisted Studies of Language and Culture*. Oxford: Blackwell.

Sullivan, K. (2009). Grammatical Constructions in Metaphoric Language. In B. Lewandowska-Tomaszczyk, & K. Dziwirek (Eds)., *STUDIES IN COGNITIVE CORPUS LINGUISTICS*, Peter Lang.

Sullivan, K. (2016). Integrating constructional semantics and conceptual metaphor. *Constructions and Frames*, *8*(2), 141-165.

Trčková, D. (2015). Representations of Ebola and its victims in liberal American newspapers. *Topics in Linguistics, 16*, 29–41.

Wang, H., Runtsova, T., & Chen, H. (2013). Economy is an organism – a comparative study of metaphor in English and Russian economic discourse. *Text & talk, 33*(2), 259–288.

Wicke, P., & Bolognesi, M.M. (2020). Framing COVID-19: How we conceptualize and discuss the pandemic on Twitter". *PLOS ONE*, 5(9), e0240010.

Wicke, P. & Bolognesi, M.M. (2021). Covid-19 Discourse on Twitter: How the Topics, Sentiments, Subjectivity, and Figurative Frames Changed Over Time. *Frontiers in Communication, 6, 651997.*

Wallis, P., & Nerlich, B. (2005). Disease metaphors in new epidemics: the UK media framing of the 2003 SARS epidemic. *Social Science & Medicine*, *60*(11), 2629–2639.

Williams, R. (1976). *Keywords: A vocabulary of culture and society*. Glasgow: Fontana/Macmillan.

Zeng, H., Burgers, C., & Ahrens, K. (2021). Framing metaphor use over time: 'Free Economy' metaphors in Hong Kong political discourse (1997-2017). *Lingua, 252*, 102955.

**Appendix A: corpus sources**

| Publications | Publication type | Publication scope |
|---|---|---|
| 27esimaora | blog | national |
| 7 (Brescia, Bergamo, Pavia) | blog | local |
| Adnkronos | magazine | national |
| AGI | blog | national |
| allthenews.altervista | blog | national |
| Avvenire | newspaper | national |
| BergamoNews | newspaper | local |
| Calciomercato.com | blog | national |
| Ciociaria Oggi | newspaper | local |
| Collettiva.it | blog | national |
| Corriere Adriatico | newspaper | local |
| Corriere dell'Umbria | newspaper | local |
| Corriere Irpinia | newspaper | local |
| Corriere Quotidiano | newspaper | national |
| Corriere Veneto | newspaper | local |
| Exibart | magazine | national |
| Euronews | newspaper | national |
| Fata Turchina Economics | blog | national |
| Fortune Italia | magazine | national |
| Gazzetta del Sud/Sicilia | newspaper | local |
| Genova Quotidiana | newspaper | local |
| Gruppo Senato Forza Italia | blog | national |
| Huffington Post | magazine | national |
| Il Corriere della Sera | newspaper | national |
| Il Corriere Fiorentino | newspaper | local |
| Il Dolomiti | newspaper | local |
| Il Fatto Quotidiano | newspaper | national |
| Il Gazzettino/nordest | newspaper | local |
| Il Giornale | newspaper | national |
| Il Giornale dello sportivo | newspaper | national |
| Il Giornale Locale | newspaper | local |
| Il Giorno dopo | blog | national |
| Il Manifesto | newspaper | national |
| Il Mattino | newspaper | local |
| Il Messaggero | newspaper | national |
| Il Post | newspaper | national |
| Il Resto del Carlino/ Bologna | newspaper | local |
| Il Resto del Carlino/ Marche | newspaper | local |
| Il Resto del Carlino | newspaper | national |
| Il Sole 24 Ore | newspaper | national |
| Il Tirreno | newspaper | local |
| Infonaples | blog | local |
| Informazione.it | blog | national |
| Internazionale | magazine | national |
| Iodonna.it | magazine | national |
| Istituto Affari Internazionali | blog | national |
| La Gazzetta del Mezzogiorno | newspaper | local |
| La Nazione | newspaper | national |
| La Nuova Sardegna | newspaper | local |
| La Repubblica | newspaper | national |
| La Sicilia | newspaper | local |
| La Stampa | newspaper | national |
| Le Scienze/espresso | magazine | national |
| Leggo.it | newspaper | national |
| Levante News | newspaper | local |
| Libero | newspaper | national |
| Metro | newspaper | local |
| Motoclub Bergamo | blog | local |
| Msn.com | blog | national |
| Next Quotidiano | newspaper | national |
| Notizie.tiscali.it | blog | national |
| Open Innovation Regione Lombardia | blog | local |
| Polesine.com | blog | local |
| Primabergamo.it | blog | local |
| Quotidiano.net | newspaper | national |
| Redattoresociale.it | blog | national |
| Reggio Online | blog | local |
| SiracusaOggi | newspaper | local |
| Start magazine | magazine | national |
| Terni Today | newspaper | local |
| The Post Internazionale | newspaper | national |
| Torino Oggi | newspaper | local |
| Trieste News | newspaper | local |
| Valori.it | blog | national |
| Wallstreetitalia.com | blog | national |

| Zero Zero News | blog | national |

**Appendix B: metaphorical networks across topics**

1. CRISES ARE NATURAL FORCES (topics: Health, Society, Economy)

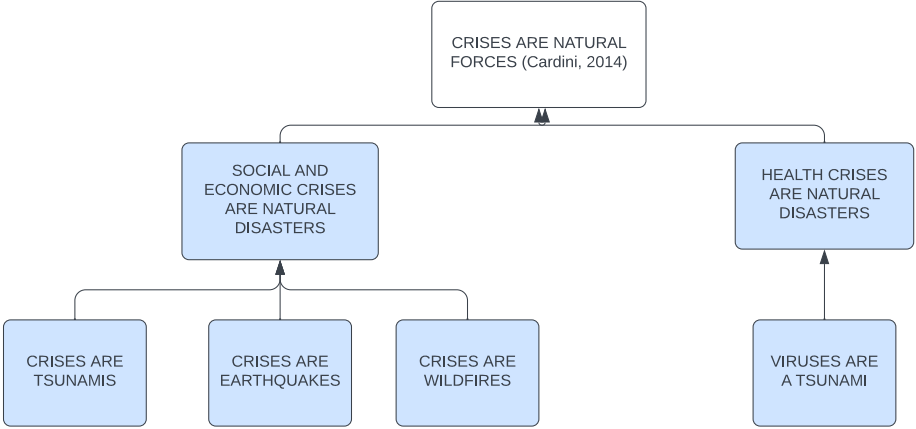

2. COMPLEX SYSTEMS ARE STRUCTURED OBJECTS (topics: Economy, Society)

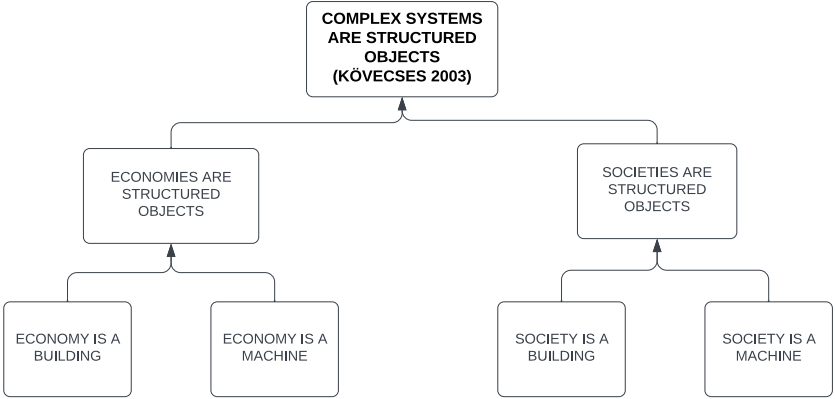

3. COMPLEX SYSTEMS ARE LIVING ORGANISMS (topics: Health, Economy, Society)

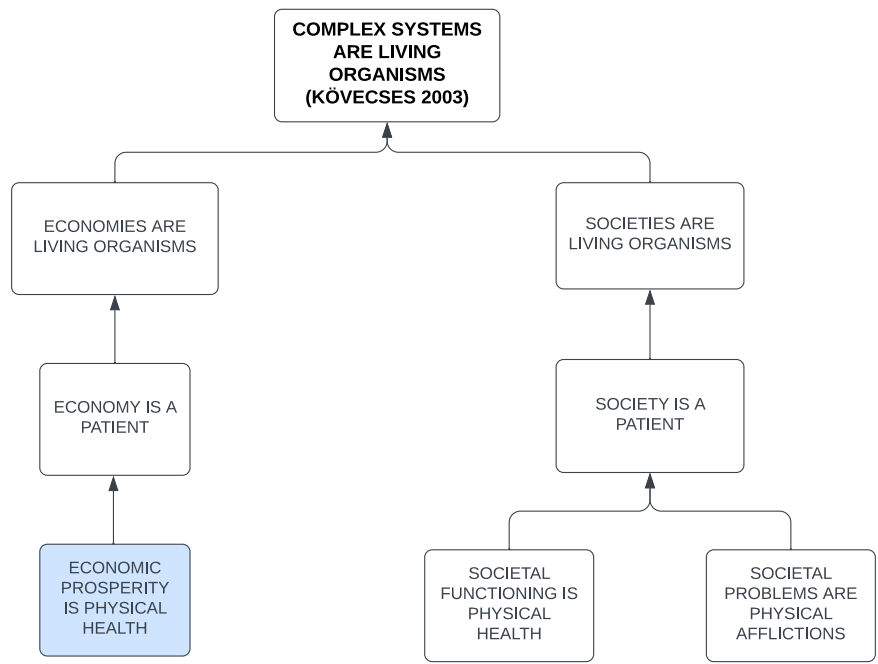